\documentclass[runningheads]{llncs}
\usepackage[T1]{fontenc}
\usepackage{graphicx}
\usepackage{hyperref}
\usepackage{siunitx}
\usepackage[capitalise]{cleveref}
\usepackage{acronym}
\usepackage{todonotes}
\usepackage{tikz}
\usetikzlibrary{shapes.geometric, arrows, shadows, positioning, fit, backgrounds, matrix}
\tikzset{%
block/.style    = {draw, thick, rectangle, anchor=west, rounded corners, align=center,
minimum height = 1em},
line/.style     = {draw, thick, -latex', shorten >=0pt},
cloud/.style    = {draw, ellipse,fill=red!20, node distance=3cm,
minimum height=2em},
container/.style = {draw, rectangle,dashed,inner sep=0.3cm, rounded corners, fill=yellow!20, node distance=3cm,
minimum height=2em},
pinstyle/.style = {pin edge={latex-, black, thick, shorten <=-1pt},
pin distance=1cm},
point/.style    = {coordinate},
bidirected/.style={latex-latex, thick},
el/.style       = {inner sep=2pt, align=left, sloped},
}
\begin{document}
%
\title{RoboCup@Home 2024 OPL Winner NimbRo: Anthropomorphic Service Robots using Foundation Models for Perception and Planning}
\titlerunning{RoboCup@Home 2024 OPL Winner NimbRo}

\acrodef{ROS}{Robot Operating System}
\acrodef{RGB-D}{Red Green Blue Depth}
\acrodef{DoF}{Degrees of Freedom}
\acrodef{FoV}{Field of View}
\acrodef{LLM}{Large Language Model}
\acrodefplural{LLM}[LLMs]{Large Language Models}
\acrodef{LiDAR}{Light Detection and Ranging}
\acrodef{GPSR}{General Purpose Service Robot}
\acrodef{EGPSR}{Enhanced General Purpose Service Robot}
\acrodef{HRI}{Human-Robot Interaction}

\newcommand{\robotname}{TIAGo++}
%

\author{Raphael Memmesheimer, Jan Nogga, Bastian Pätzold, Evgenii Kruzhkov, Simon Bultmann, Michael Schreiber, Jonas Bode, Bertan Karacora,\\ Juhui Park, Alena Savinykh, and Sven Behnke} 

\institute{Autonomous Intelligent Systems, Computer Science Institute VI, Lamarr Institute for Machine Learning and Artificial Intelligence, and Center for Robotics,\\University of Bonn, Germany\\
\url{https://www.ais.uni-bonn.de/nimbro/@Home/}
}

\authorrunning{R. Memmesheimer et al.}
%
\maketitle              
\begin{abstract}
We present the approaches and contributions of the winning team NimbRo@Home at the RoboCup@Home 2024 competition in the Open Platform League held in Eindhoven, NL.
Further, we describe our hardware setup and give an overview of the results for the task stages and the final demonstration.
For this year's competition, we put a special emphasis on open-vocabulary object segmentation and grasping approaches that overcome the labeling overhead of supervised vision approaches, commonly used in RoboCup@Home.
We successfully demonstrated that we can segment and grasp non-labeled objects by text descriptions.
Further, we extensively employed \acp{LLM} for natural language understanding and task planning.
Throughout the competition, our approaches showed robustness and generalization capabilities.
A video of our performance can be found online\footnote{\url{https://www.ais.uni-bonn.de/videos/RoboCup_2024}}. 
\end{abstract}

\section{Introduction}
\label{sec:introduction}

\begin{figure}[ht]
	\centering
	\includegraphics[width=.8\linewidth]{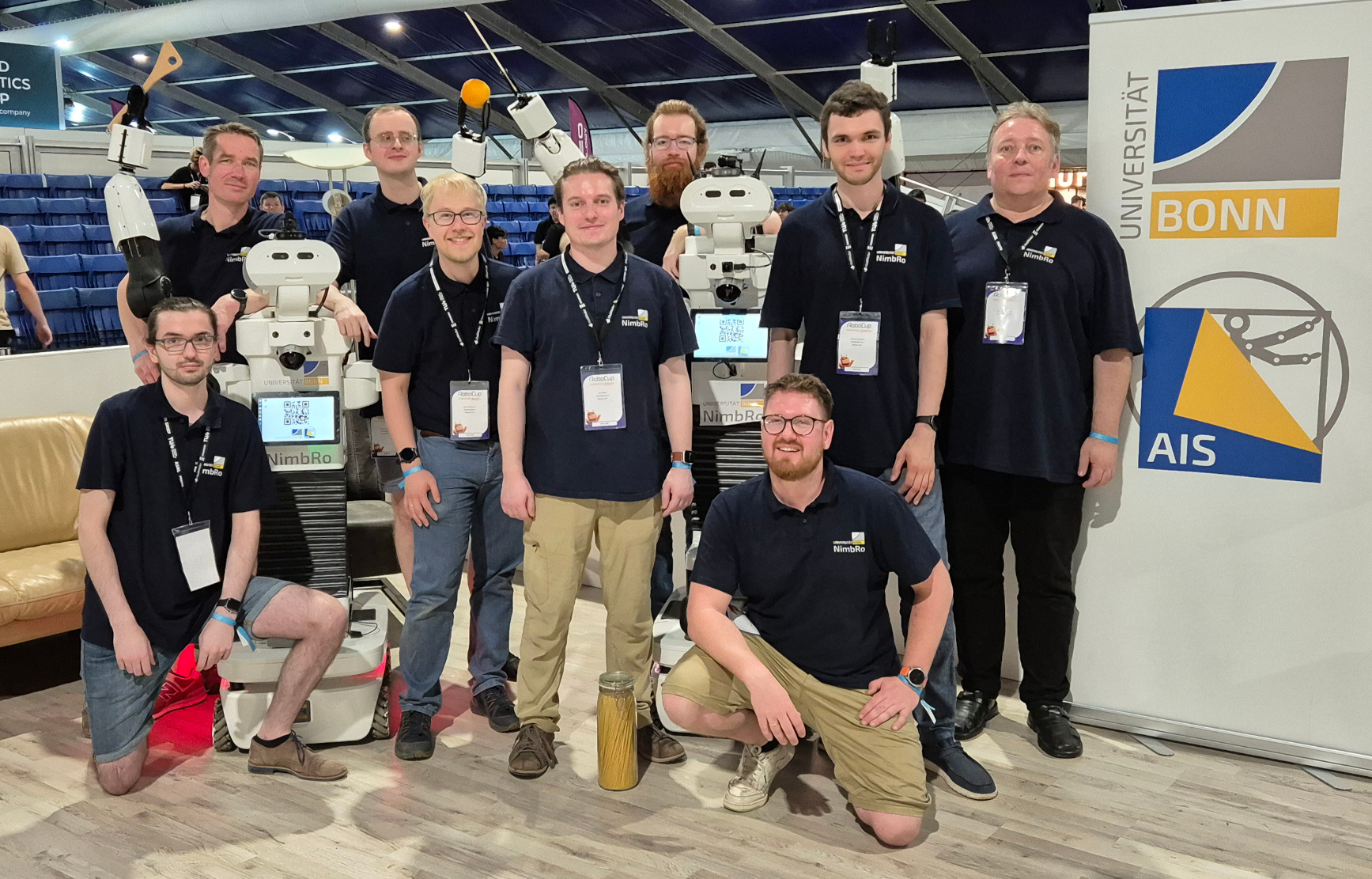}
	\vspace*{-1ex}
	\caption{The NimbRo@Home team at RoboCup 2024 in Eindhoven, NL.}
	\vspace*{-2ex}
	\label{fig:team}
\end{figure}

NimbRo has a well-established track record of successful participation in robotic competitions ranging from humanoid soccer in the RoboCup AdultSize class~\cite{Pavlichenko:Winner2024}, mobile manipulation in unstructured environments at the DLR SpaceBot Cup~\cite{SchwarzBDSPLSB:Frontiers16} and the DARPA Robotics Challenge~\cite{schwarz2017nimbro}, cluttered bin picking  at the Amazon Picking\,/\,Robotics Challenges~\cite{schwarz2017nimbropicking,SchwarzLGKPSB:ICRA18}, multiple UGV and UAV tasks at MBZIRC 2017~\cite{SchwarzDLPPRRSS:JFR19,BeulNQRHPHB:JFR19} and 2020~\cite{LenzQPRRSSSSB:FR22,BeulSQSBSRPRLSSSB:FR22}, and immersive telepresence at the ANA Avatar XPRIZE competition~\cite{lenz2023nimbro}.
After winning RoboCup@Home three times in a row 2011-2013~\cite{StucklerSB:Frontiers16}, we participated again at RoboCup@Home 2023 in Bordeaux~\cite{memmesheimernimbrotdp2023} and now won the RoboCup@Home 2024 competition in the Open Platform League. \Cref{fig:team} shows our team at RoboCup 2024 in Eindhoven, NL. 

In this paper, we give an overview of the hardware setup, describe the approaches and methods used in the competition and present the results of the RoboCup 2024 world championship in the @Home Open Platform League.
Our team won the competition with a final score of 8,852, followed by team Tidyboy-OPL (South Korea) with a score of 7,495 and SocRob@Home (Portugal) with a score of 6,901.
Novelties of this year's participation include a modified robot platform, improvements in our grasping and vision approaches and the integration of \acp{LLM} for task planning.
We focused on the integration of open-vocabulary approaches for object segmentation.

\begin{figure}[ht]
    \centering
    \includegraphics[width=.69\linewidth]{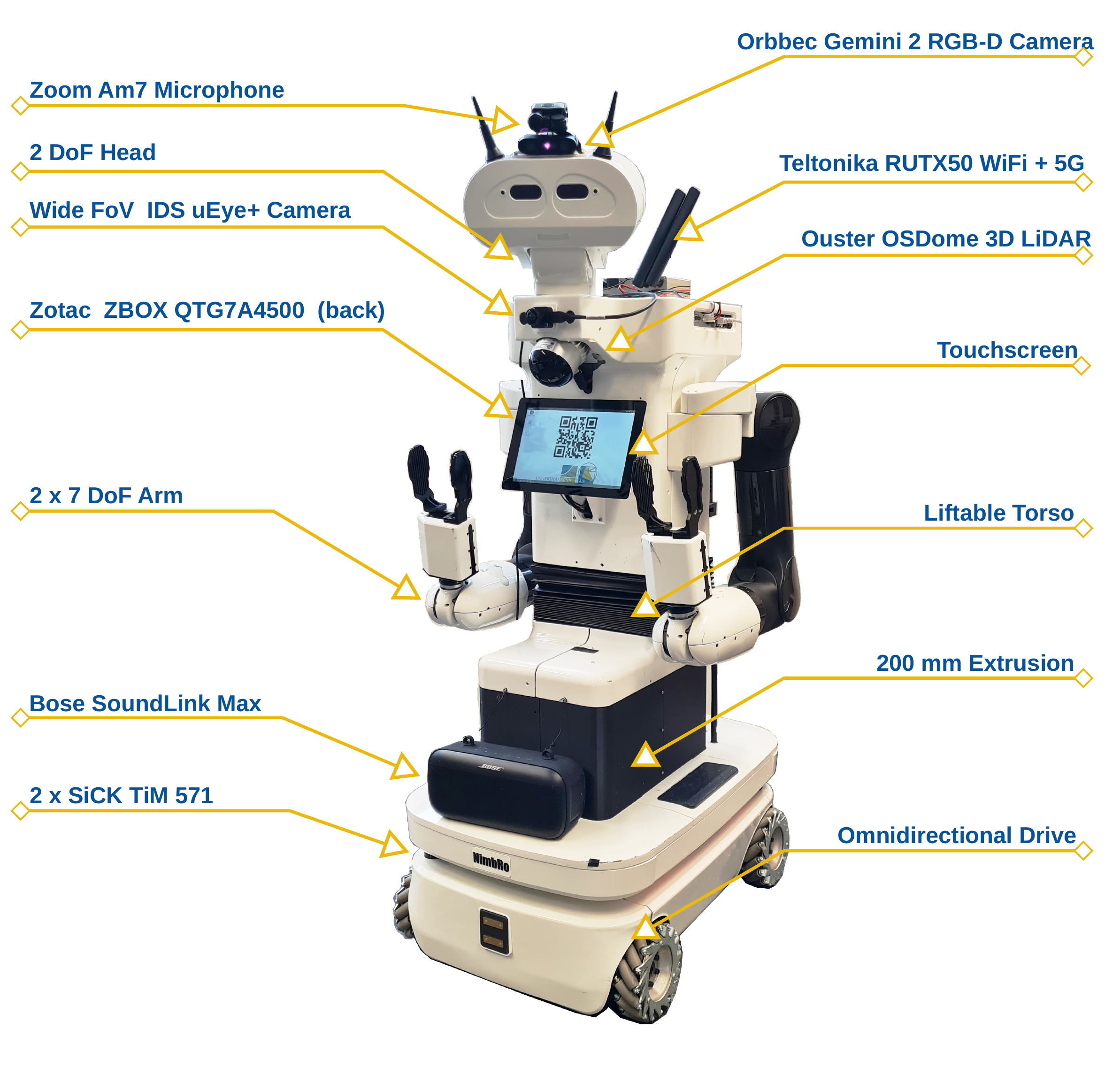}
		\vspace*{-4ex}
    \caption{Enhanced \robotname{} omnidirectional robot platform.}
 	\vspace*{-2ex}
   \label{fig:robot}
\end{figure}
\section{Hardware}

For RoboCup@Home 2024 participation we used a modified \robotname{} robot~\cite{pages2016tiago} (see \Cref{fig:robot}) which is equipped with an omnidirectional mobile base, a linear liftable torso with two 7-\ac{DoF} arms and a pan-tilt-unit with an RGB-D camera.
A ZBOX QTG7A4500 with an NVIDIA RTX A4500 is used for model inference and is mounted on the robot's back.
An Ouster OSDOME-128 with a \ang{180} FOV is used to gather a  frontal hemispherical view of the robot for small obstacle avoidance and precise estimates for distant 3D perception.
The \ac{LiDAR} is calibrated against a wide FoV IDS uEye+ Camera to colorize the point cloud and to project detections from the camera to the \ac{LiDAR} frame.
For 2D mapping and localization, two Sick TiM 571 \ac{LiDAR} sensors are used to cover the robot's surroundings.
A 10-inch IPS touch screen is mounted at the front of the robot to improve human-robot interaction.
For speech recognition, a Zoom Am7 microphone is used.
A Teltonika RUTX50 Wi-Fi + 5G router served as a hybrid Wi-Fi and 5G connection for the robot, using the 5G connection when it couldn't connect to the Wi-Fi network to ensure connectivity to online API services we utilize.
We increased the height of the robot by \SI{200}{mm} to improve the grasping workspace and limit the extent of the elbow joints when grasping objects from table and shelf heights.
The speaker of the robot was upgraded to Bose SoundLink Max to enhance its audibility in noisy environments.
We use ROS 2 Foxy~\cite{macenski2022robot} on the robot's back computer and bridge ROS 1 topics from the robot's onboard computer.
A second robot with a similar configuration was used for redundancy and to allow for parallel testing.

\section{Software for Perception and Planning}
\label{sec:approaches}

The developed software modules address mapping and navigation, person and object perception, human-robot interaction, and task planning using \acp{LLM}.

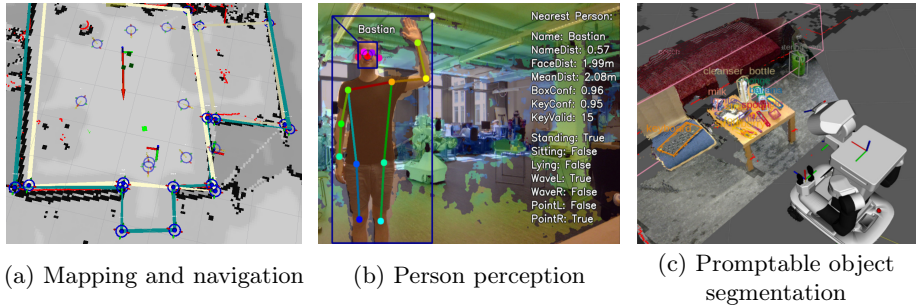
\begin{figure}[t]
  \centering
  \begin{tikzpicture}[node distance = 0cm]
    \node[] at (0,0) (a) {\includegraphics[height=3.2cm]{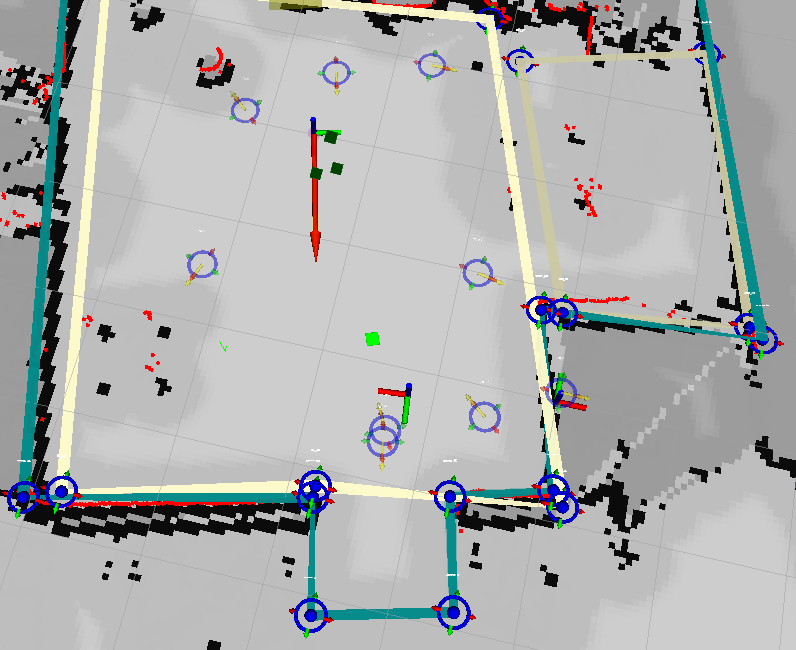}};
    \node[right=of a] (b) {\includegraphics[height=3.2cm, width=4cm]{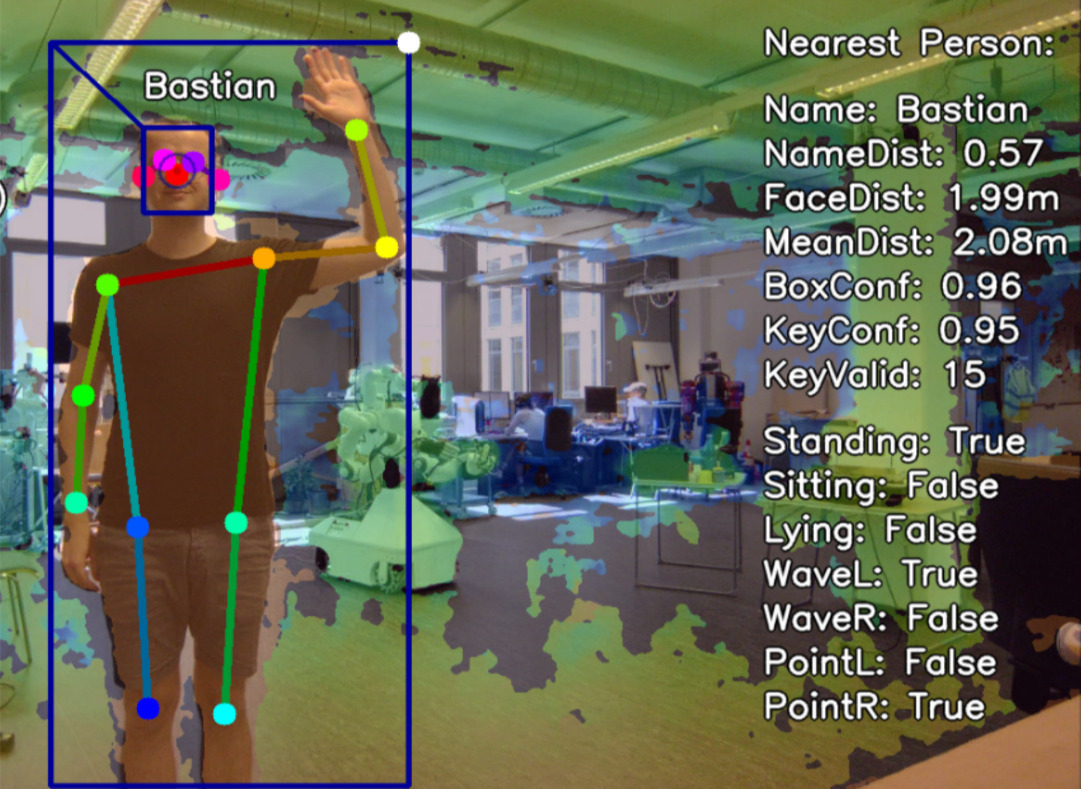}};
    \node[right=of b] (c) {\includegraphics[height=3.2cm, width=3.7cm]{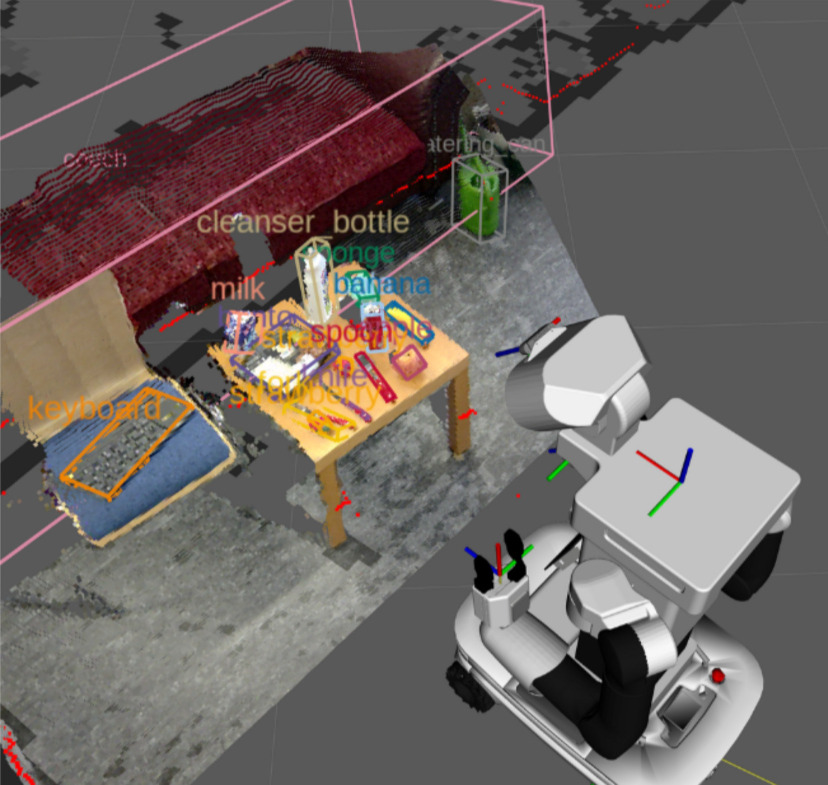}};
    \node[below=of a.south,align=center,yshift=-.05cm] (a_caption) {(a) Mapping and navigation};
    \node[below=of b.south,align=center,yshift=-.05cm] (b_caption) {(b) Person perception};
    \node[below=of c.south,align=center,yshift=+.1cm] (c_caption) {(c) Promptable object\\ segmentation};
  \end{tikzpicture}
  \vspace{-.8cm}
  \caption{Software modules used in the competition. (a) Map including location markers and annotated regions. (b) Person detection using YOLO V8, body pose estimation, action and face recognition. (c) Projected object segments using mmGrounding-DINO.}
  \vspace{-.2cm}
  \label{fig:approaches}
\end{figure}

\subsection{Mapping and Navigation}
\label{sec:mapping_navigation}

We employ the SLAM Toolbox~\cite{macenski2021slam} to perform the mapping and utilize AMCL for localization. SLAM toolbox is a graph-based approach with high mapping accuracy due to loop closure support.
When the operation area is unknown in advance (Carry My Luggage task), we use the SLAM Toolbox localization mode that updates the graph to new regions.
In known environments, we utilize pose location markers to encode poses of interest. In addition, we can use those markers to define regions that can be used to distinguish between different areas (rooms), to constrain person and object search areas, or to restrict specified areas for the robot. 
An example of the underlying map is shown in \Cref{fig:approaches}\,a. We edit the map in a web interface created with Viser Studio.

\subsection{Person Perception}
\label{sec:person_detection}

The exemplary person detection shown in \Cref{fig:approaches}\,b uses YOLO V8~\cite{Jocher_Ultralytics_YOLO_2023} models for both pose and face detection.
Both methods are combined by computing the Mean Intersection over Union (MIoU) of the bounding boxes of the face detection and the estimated face from the pose.
For face recognition, we employ a VGG-Face~\cite{vgg-face} embedding model from DeepFace~\cite{serengil2024lightface}.
Compatible with both the Gemini and IDS cameras, the system can access depth for each detection, allowing it to naively recognize gestures such as hand waving, and discard detections of persons outside a specified area, such as the competition arena.

For tracking, we employed a combination of 2D person detector from an RGB-D camera and a re-identification approach~\cite{DBLP:conf/fusion/WojkeMP17} that continuously collects features of the tracked person to distinguish the operator from the remaining crowd.
In case the operator is lost, the re-identification approach is used to re-identify the operator.
If that fails, the robot asks the operator to wave to the robot to reinitiate the tracking.
During the tracking phase, we adjust the robot’s head to continuously look at the operator. This improves the interaction and should minimize that the operator's track is lost when the robot's navigation is not optimally facing the operator.

\subsection{Object Perception}
\label{sec:object_segmentation}

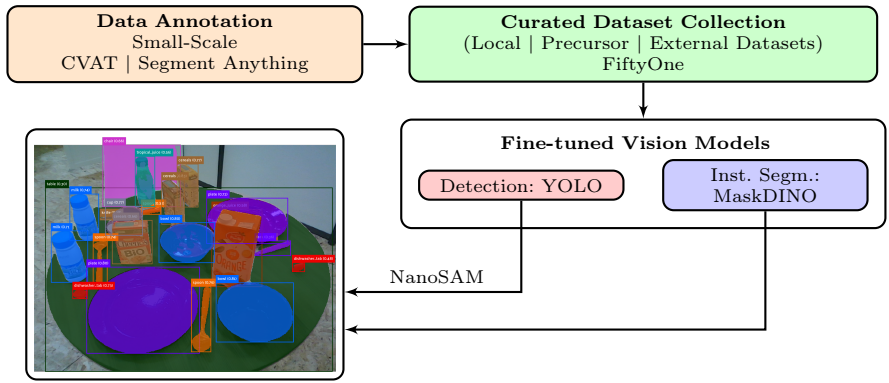
\begin{figure*}[t]
  \centering
  \begin{tikzpicture}[node distance = 0.6cm, auto, every node/.style={font=\scriptsize}]
    \node [block, text width=4.5cm, fill=orange!20] (data) {\textbf{Data Annotation} \\ Small-Scale \\ CVAT | Segment Anything};
    \node [block, right=of data, text width=6.0cm, fill=green!20] (curated) {\textbf{Curated Dataset Collection} \\ (Local | Precursor | External Datasets)\\ FiftyOne};

    \node [below=of curated, text width=3.8cm] (caption) {\bf Fine-tuned Vision Models};
    \node [matrix, below=of caption, yshift=0.7cm, column sep=0.5cm] (models) {
      \node [block, text width=2.5cm, fill=red!20] (yolo) {Detection: YOLO}; &
      \node [block, text width=2.5cm, fill=blue!20] (maskdino) {Inst. Segm.: MaskDINO}; \\
    };

    \node [block, fit={(caption) (models)} ] (models) {};

    \node [block, below=of data] (result) {\includegraphics[width=4cm]{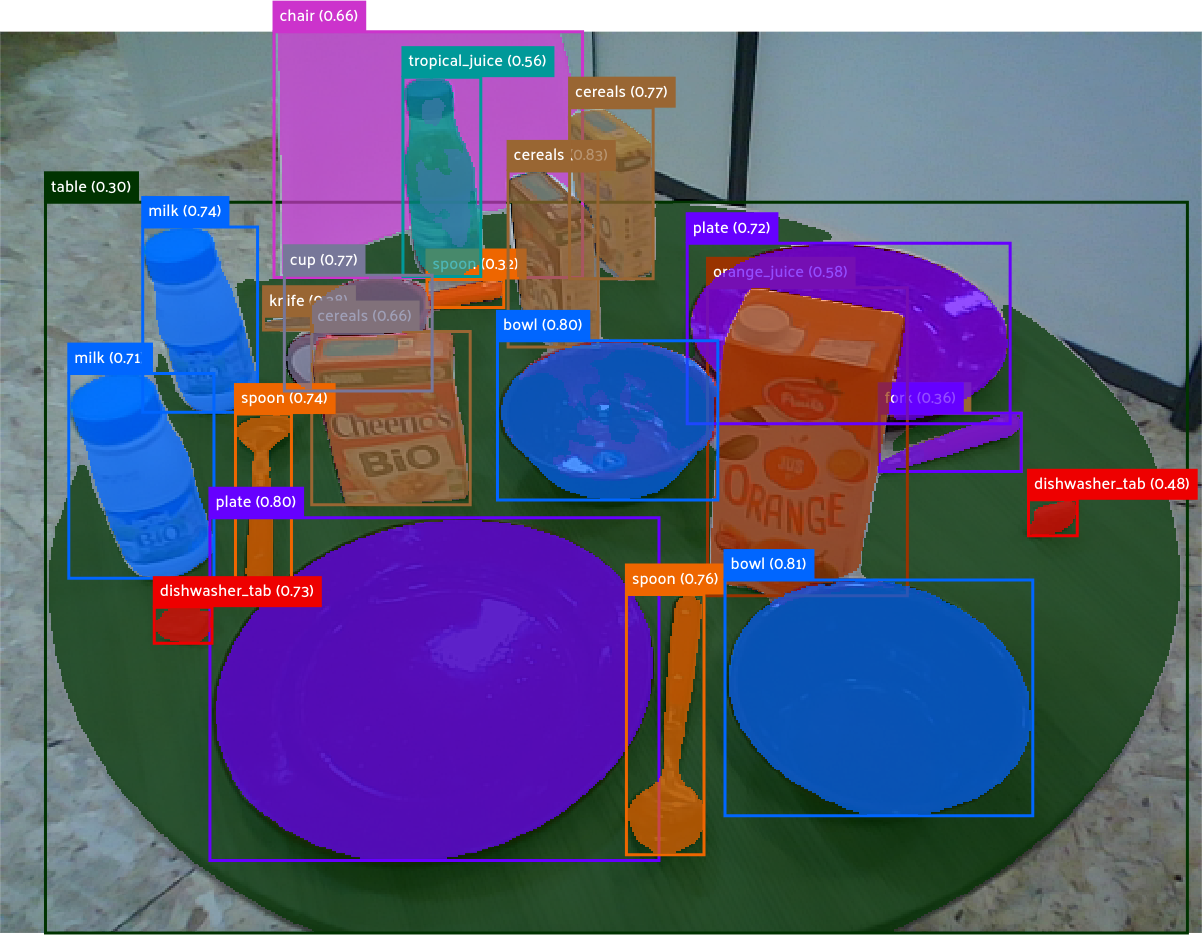}};
    \path [line] (data) -- (curated);
    \path [line] (curated) -- (models);
    \path [line] (yolo.south) |- node [text width=2.5cm,midway,above,xshift=-0.5cm] {NanoSAM} ([xshift=0.5cm, yshift=-0.5cm]result.east) -- ([yshift=-0.5cm]result.east);
    \path [line] (maskdino.south) |- ([xshift=0.5cm, yshift=-1cm]result.east) -- ([yshift=-1cm]result.east); %
  \end{tikzpicture}
  \caption{Object perception pipeline including annotation, curation, and backbone models. We annotate data semi-automatically using CVAT and Segment Anything. This data is then curated using FiftyOne and combined with external datasets. The curated data is used to fine-tune YOLO and MaskDINO models. Additionally, we employ NanoSAM to recover segments from YOLO detections.}
  \vspace{-.2cm}
  \label{fig:vision_pipeline}
\end{figure*}

In this year's RoboCup participation we utilized both, supervised and open-vocabulary object segmentation approaches and could flexibly switch between the two approaches.
The closed vocabulary training and detection pipeline is supported by a dataset of household objects spanning 63,516 object instances from 104 classes in 5,088 frames. It was captured in our lab in Bonn as well as at previous competitions and annotated using SAM~\cite{kirillov2023segment} interactively. At RoboCup, the dataset was extended to include the local objects in cluttered arrangements. Over the setup days, more data was collected for objects which were not performing well in the detection metrics and for difficult scenarios such as faraway objects on the floor. These local datasets can be mixed with the base dataset by remapping labels to visually similar local object classes to improve recall or mixing in unrelated objects as negative examples to improve precision. Our object perception pipeline is depicted in \Cref{fig:vision_pipeline}. Fine-tuning of detection (YOLO~\cite{Jocher_Ultralytics_YOLO_2023}) or instance segmentation (MaskDINO~\cite{li2023mask}) models is done on a task per task basis, targeting only the object classes required in that task. Our base models are pretrained on large image datasets (COCO, OpenImages), to assure that the feature extraction layers are performant from the start. In this setup, fine-tuning a task-specific model on a single RTX 4090 takes between 10 and 30 minutes. In Eindhoven, we used MaskDINO models rather than YOLO because they proved more robust to the changes in lighting given our training data.

While open-vocabulary detectors are not yet as performant in terms of inference speed and detection metrics as their closed vocabulary counterparts, we find mmGrounding-DINO~\cite{zhao2024open} to suffice for many household tasks. In this year's RoboCup vision strategy, we used this model when there were few relevant object classes (Serving Breakfast, Clean the Table). The prompts were manually designed using our locally captured dataset for evaluation and could be changed on the fly during the task depending on its state, such as when checking for chairs, searching for cutlery and tableware or looking for the dishwasher tab in  the Clean the Table task, respectively. Since Grounding-DINO models output bounding boxes, we use these as prompts in NanoSAM to obtain the instance segmentations we require for grasping.

\begin{figure}[t]
    \centering
    \begin{tikzpicture}[node distance = 0cm, auto, every node/.style={font=\small}]
      \node[] (a) at (0,0) {\includegraphics[height=0.4\linewidth]{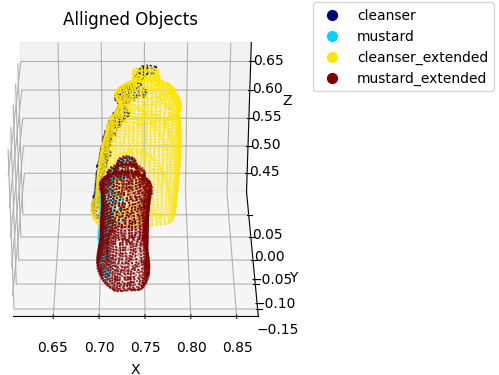}};
      \node[right=of a] (b) {\includegraphics[height=0.4\linewidth]{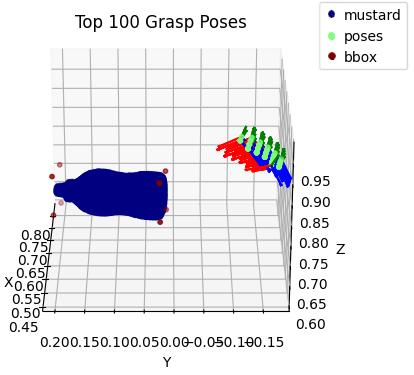}};
      \node[below=of a.south,align=center,yshift=-.0cm] (a_caption) {\hspace*{-10ex}(a) Object registration};
      \node[below=of b.south,align=center,yshift=-.0cm] (b_caption) {(b) Grasp proposals};
    \end{tikzpicture}
    \vspace{-.7cm}
    \caption{3D object perception and grasping. (a) If available, 3D models can be registered to partial point clouds of all detected objects. (b) The grasp proposals minimizing the grasping cost function for approaching a bottle of mustard lying on a surface.}
    \label{fig:obj_registration_grasp_proposals}
\vspace*{-2ex}		
\end{figure}

\subsection{Grasping}
\label{sec:grasping}

For manipulation, the robot is aligned based on the orientation of the object and a reachability map precomputed using cuRobo~\cite{sundaralingam2023curobo}. Masking the object in the depth image yields a partial point cloud, which is completed by registration to a 3D model if one is available for the corresponding class label. The final point cloud is then approximated by a 3D oriented bounding box. 

Next, potential grasp poses are sampled on a quadrant sphere covering the side of the object that is facing the robot, their base orientations placing the gripper opening towards the object CoM with the fingers oriented vertically. These base poses are augmented by twist angles around the forward axis in the end effector frame. 

In a subsequent filtering step, grasp poses which are in collision with the environment are identified using a KD-Tree lookup in a composite point cloud from the RGB-D camera and 3D LiDAR. This check is repeated for the approach trajectories of the remaining candidate poses. Then, the object 3D oriented bounding box is transformed into the frame of each grasp pose. Poses in which the object width exceeds the gripper width are removed. From the remaining poses, the most affordable one is selected for execution using a heuristic which considers the availability of similar precomputed pre-grasp poses which are not in collision with the environment, the distance from obstacles and the margin between grasp pose and the borders of the robots' workspace. Figure \ref{fig:obj_registration_grasp_proposals}\,b shows example grasp proposals.

\subsection{Speech Synthesis and Recognition}
\label{sec:speech}

The foundation of our audio processing pipeline is the JACK Audio Connection Kit~\cite{JACK}, which provides capabilities for real-time audio processing and interfacing to connected audio hardware.
To cope with challenging acoustic conditions in downstream tasks, the microphone signal is pre-processed using the NVIDIA Maxine toolkit~\cite{Maxine}, which applies denoising and dereverberation to isolate clean foreground speech from background noise.
To retrieve speech commands at specific times during task execution, we use a voice activity detection model~\cite{vad} to determine beginning and end-of-speech boundaries.
Speech segments are then passed directly to a local instance of Faster Whisper~\cite{fasterwhisper} for speech recognition, implementing Radford et al.~\cite{radford2023robust}, which is capable of transcribing 99 different languages and translating them into English.
Thus, our speech recognition pipeline can be characterized as robust, grammar-free, and multilingual.

For text-to-speech synthesis, we utilize the Coqui.ai library~\cite{Eren_Coqui_TTS_2021}, which implements the end-to-end approach of Jaehyeon et al.~\cite{tts}.
We embed this model between custom pre- and post-processing modules for text normalization of numerals and punctuation, as well as loudness normalization between passes and loudness maximization to cut through loud environmental noises.

\subsection{Task Planning using Large Language Models}
\label{sec:task_planning}

To allow the robot to understand and act on complex natural language instructions where naive keyword searching is not sufficient, we use \acp{LLM}, i.e. \textit{GPT-4o}~\cite{openai2023gpt4}, which is accessed via our 5G router.
We aimed for a general solution that includes both the \textit{\ac{GPSR}} and \textit{\ac{EGPSR}} tasks provided by non-expert operators.

Our approach is based on encapsulating the robot's capabilities, such as \textit{communicating with a person}, \textit{driving to a location}, or \textit{grasping an object}, in a general way, so that a successful execution of a typical command can be achieved by calling about 3 to 15 functions.
In order for the \ac{LLM} to keep track of progress and possibly react to unexpected situations, it is important that these functions are not only general and robust, but also provide appropriate textual feedback.
The \ac{LLM} is then used to execute one function after another until the command is accomplished, using the \textit{function calling} feature provided by many SOTA models.
The \ac{LLM} can determine that the command has been accomplished by calling an appropriate function that advances the state machine of the underlying task.
We have also worked on analyzing the performance of different prompting methods in a similar context~\cite{bode2024prompting}.
In addition, we use \acp{LLM} to reject commands that require capabilities that the robot either does not have or that take a long time to execute, so that moving on to the next task increases the overall score.

\section{Competition Results}
\label{sec:results}

We briefly present the results of the predefined tasks and the final demonstration of the RoboCup@Home 2024 competition.
For complete task descriptions, we refer to the rule book~\cite{rulebook_2024}.
In this year's competition, the tests were executed twice in different arenas.

\subsection{Stage~1}
\label{sec:stage1}

\begin{figure}[t]
  \centering
  \begin{tikzpicture}[node distance = 0cm]
    \node[] (a) at (0,0) {\includegraphics[height=3.8cm]{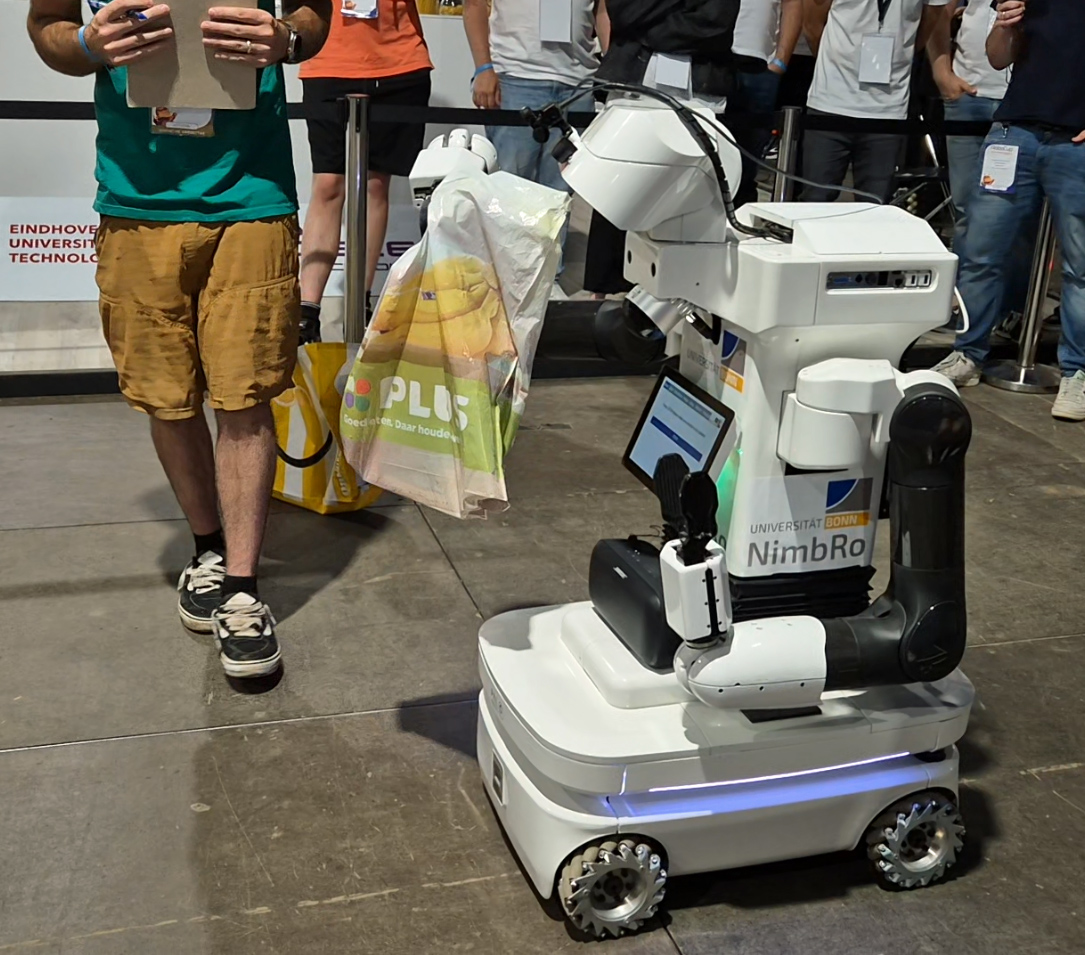}};
    \node[right=of a] (b) {\includegraphics[height=3.8cm]{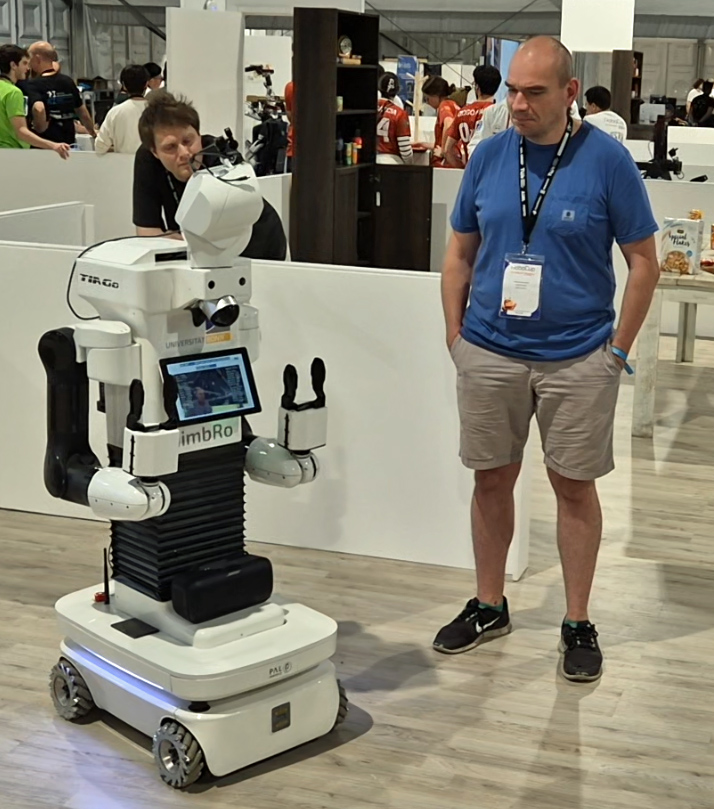}};
    \node[right=of b] (c) {\includegraphics[height=3.8cm]{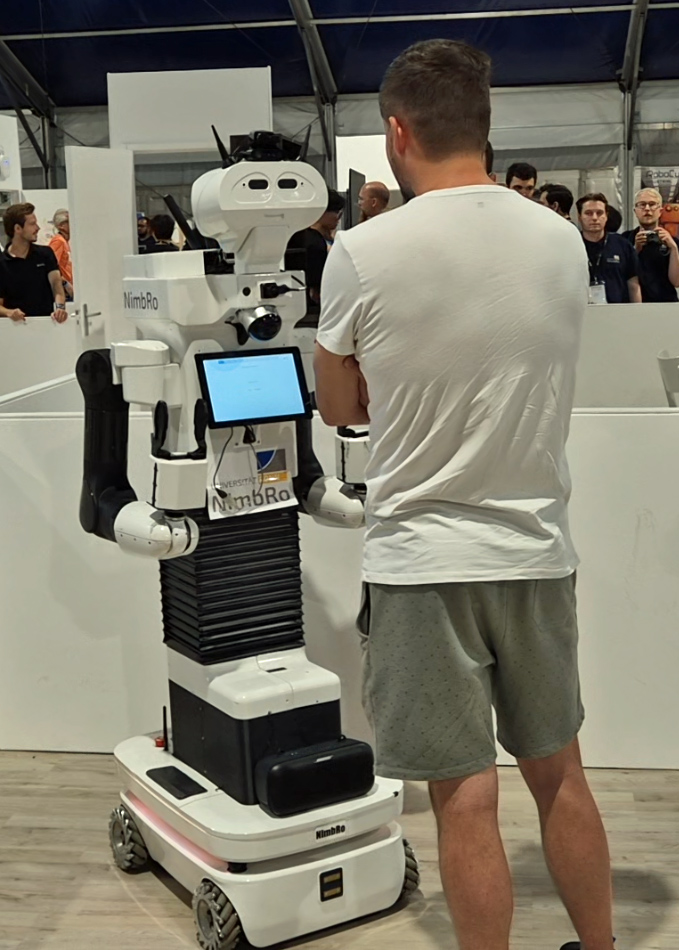}};
    \node[below=of a.south,align=center,yshift=+.05cm] (a_caption) {(a) Carry My Luggage};
    \node[below=of b.south,align=center,yshift=+.05cm] (b_caption) {(b) Receptionist};
    \node[below=of c.south,align=center,yshift=+.05cm] (c_caption) {(c) \ac{GPSR}};
  \end{tikzpicture}
  \vspace{-.4cm}
  \caption{Impressions from Stage~1 tests.} 
	\vspace*{-2ex}
  \label{fig:stage1}
\end{figure}
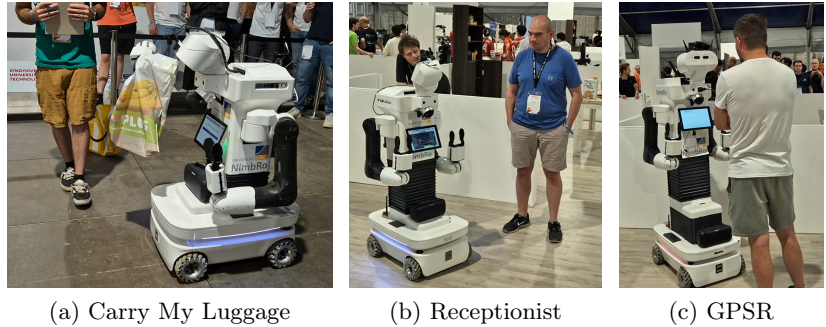

Stage~1 consists of the tasks \textit{Carry My Luggage}, \textit{Receptionist}, \textit{Serve Breakfast}, \textit{Storing Groceries} and \textit{\ac{GPSR}}.
The task duration is limited to 5\,min for all tasks.

\noindent$\circ$~In \textit{Carry My Luggage} (\Cref{fig:stage1}\,a), our robot successfully detected the correct bag to carry, picked it up successfully, memorized the operator and followed the operator while passing a crowd, a tiny object and a hard-to-see object (chair). Our robot lost track of the operator when it was avoiding the hard-to-see object.

\noindent$\circ$~In \textit{Receptionist} (\Cref{fig:stage1}\,b), our robot welcomed two guests, identified their names and favorite drinks via speech, estimated distinguishable attributes by face- and pose analysis to describe the guests to the other guests and offered free seats.
It was limited in time for offering the second seat.

\noindent$\circ$~A low score in \textit{Storing Groceries} was caused by a hardware defect which prevented the robot from utilizing the arms properly.

\noindent$\circ$~For the \textit{Serve Breakfast}, task our robot successfully grasped a spoon and a bowl, placed the bowl, and placed the spoon next to it. Finally, it perceived the cereals.
\noindent$\circ$~In \textit{\ac{GPSR}} (\Cref{fig:stage1}\,c), our robot received two commands by a naive operator: "Tell me what the smallest object on the kitchen counter is." and "Navigate to the office and hand me a pear." The robot successfully executed the first command and was limited in time for the second command. This performance yielded the highest score among all sub-leagues in the \textit{\ac{GPSR}} task.

After Stage~1, our team was ranked second with a score of 2,128 --- behind Tidyboy-OPL (2,717).

\subsection{Stage~2}
\label{sec:stage2}

Stage~2 consists of the tasks \textit{Clean the Table}, \textit{E\ac{GPSR}}, \textit{Restaurant} and \textit{Stickler for the Rules}.
The duration for the tasks is 10\,min, while the \textit{Restaurant} task is limited to 15\,min.

\begin{figure}[t]
  \centering
  \begin{tikzpicture}[node distance = 0cm]
    \node[] at (0,0) (a) {\hspace*{-1ex}\includegraphics[height=3.07cm]{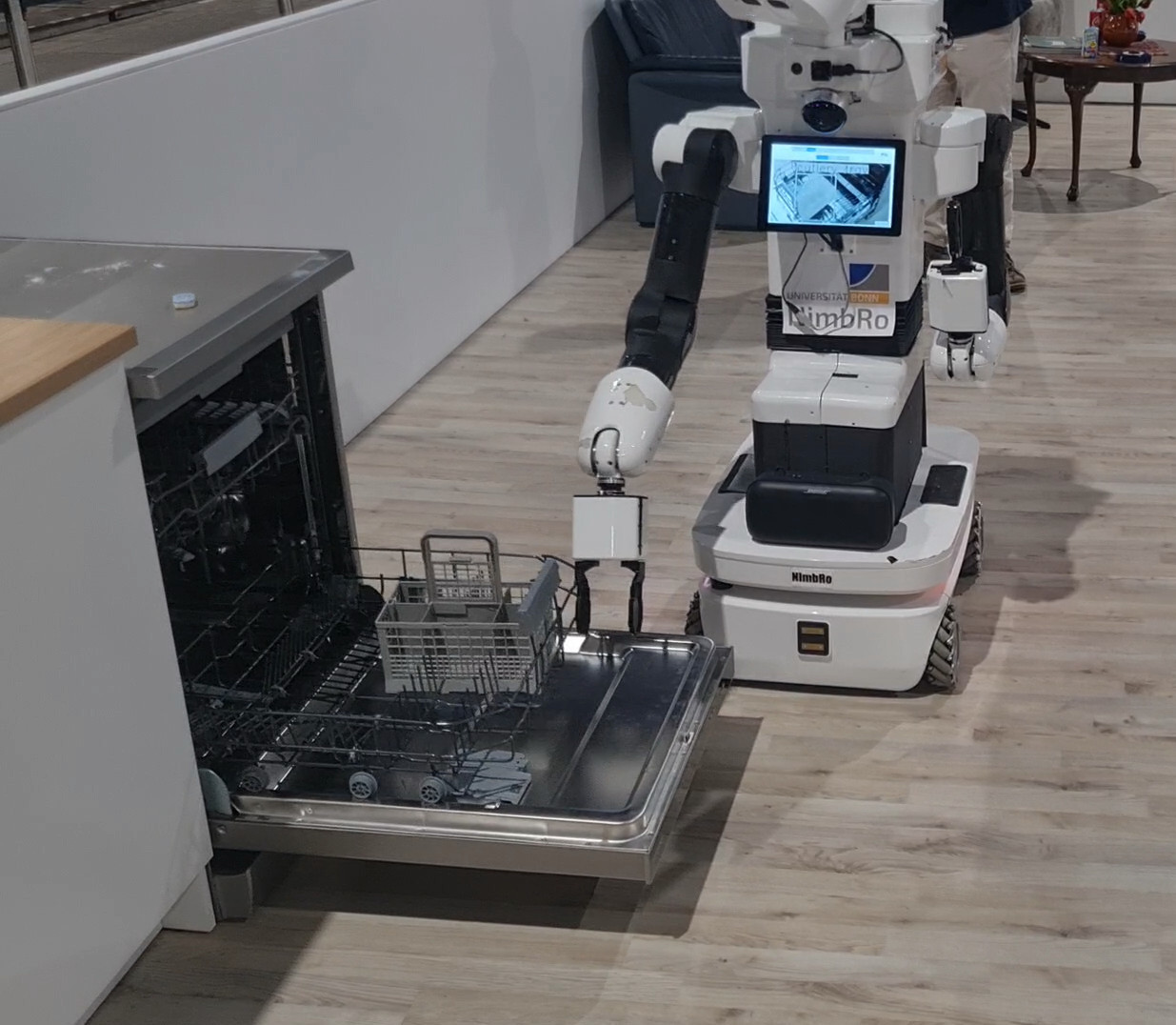}};
    \node[right=of a] (b) {\includegraphics[height=3.07cm]{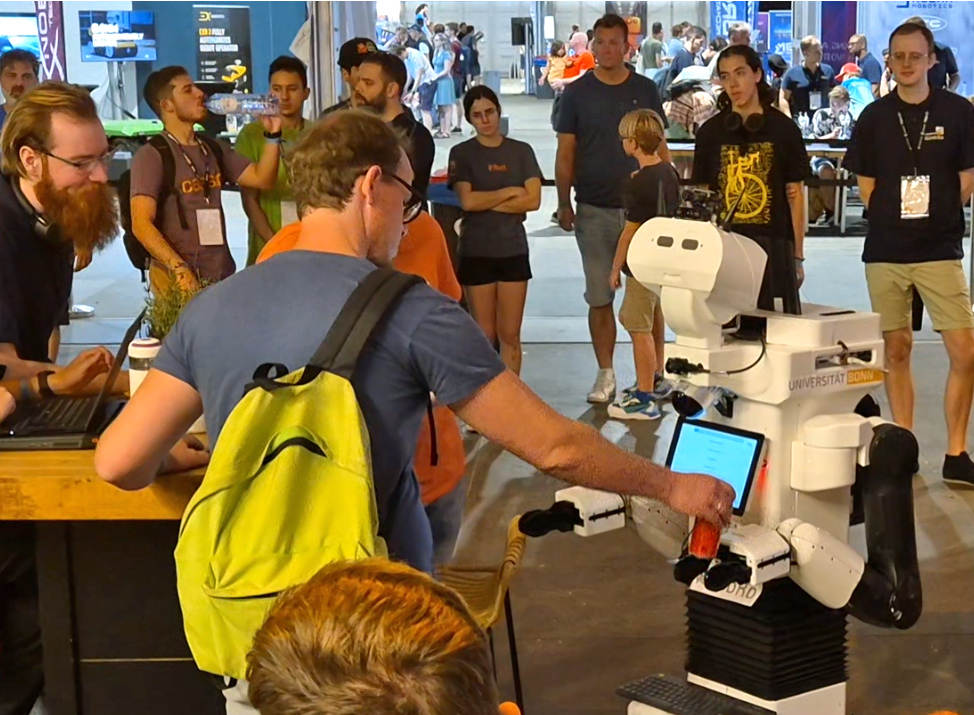}};
    \node[right=of b] (c) {\includegraphics[height=3.07cm]{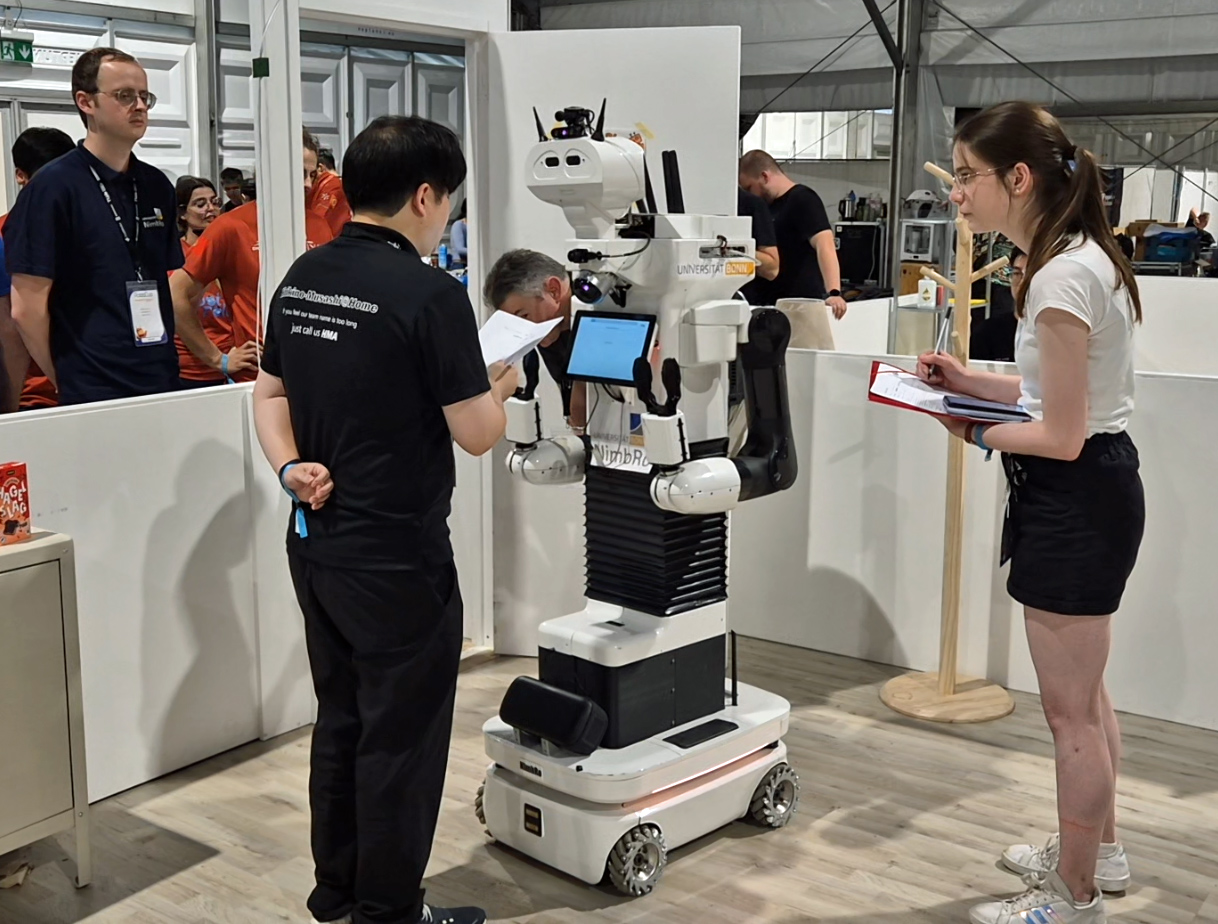}};
    \node[below=of a.south,align=center,yshift=+.05cm] (a_caption) {(a) Clean the Table};
    \node[below=of b.south,align=center,yshift=+.05cm] (b_caption) {(b) Restaurant};
    \node[below=of c.south,align=center,yshift=+.05cm] (c_caption) {(c) \ac{EGPSR}};
  \end{tikzpicture}
  \vspace{-.8cm}
  \caption{Impressions from Stage~2 tests.}
  \label{fig:stage2}
	\vspace*{-2ex}
\end{figure}
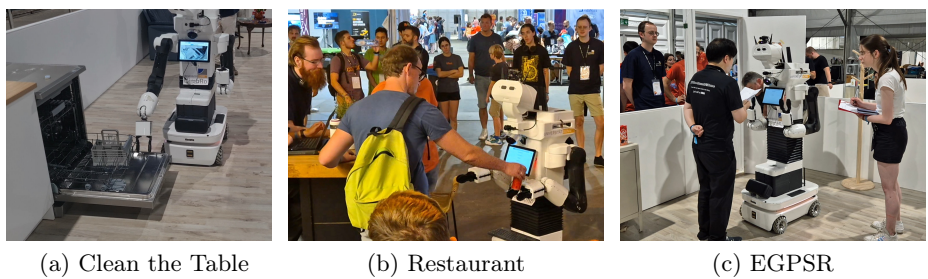

\noindent$\circ$~In \textit{Clean the Table} (\Cref{fig:stage2}\,a), our robot successfully grasped and placed a cutlery item conveniently in the dishwasher and pushed in the rack.

\noindent$\circ$~In \textit{Restaurant} (\Cref{fig:stage2}\,b), our robot detected waving guests and navigated through the unknown environment towards them.
When reaching customers, the robot took their orders using speech recognition. If environmental noise proved too challenging, the robot fell back on a touchscreen interface.
Following this pattern, our robot took orders from two different customers.
Communicating with the barman, the robot then picked up the ordered items and delivered them to the recipients.
In addition, the robot was also able to interact with detected waving guests who did not want to order anything, returning to the loop of searching for new customers and handling their orders.
We received the \textit{Waitress Captain (Best in Restaurant Test)} award as our team scored highest in the Restaurant task among all sub-leagues.

\noindent$\circ$~In \textit{\ac{EGPSR}} (\Cref{fig:stage2}\,c), our robot followed a pre-designed path, scanning for trash with a closed-vocabulary detector and instructing the judge to dispose detected objects.
Using the wide-angle IDS camera, it identified all waving persons, collected tasks, and executed them sequentially, starting with the closest non-rejected task, following our Stage~1 \textit{GPSR} approach. In the first run, we hit the time limit during the initial task. In the second run, our robot faced issues receiving a task from a quiet-voiced operator due to nearby robot band noise.

\noindent$\circ$~For the \textit{Stickler for the Rules} task, we employed open-vocabulary detection models together with a pre-trained one.
The pre-trained model detected known floor objects. To minimize false positives, training data was collected in all arena rooms with varying backgrounds. Open-vocabulary models detected shoes, socks, and drinks --- allowing for unknown item detection. Using the same model saved GPU resources. The robot patrolled the arena, stopping to detect rule offenders, managed by a custom state machine. In unexpected situations, it returned to the main patrol cycle. 

After Stage~2, we led the competition with a score of 4,918, followed by SocRob@Home (4,247) and Tidyboy-OPL (4,213).

\subsection{Final Demonstration}
\label{sec:final_demonstration}

\begin{figure}[t]
    \centering
    \begin{tikzpicture}[node distance = 0cm]
      \node[] (a) at (0,0) {\hspace*{-1ex}\includegraphics[height=2.8cm, width=3.8cm]{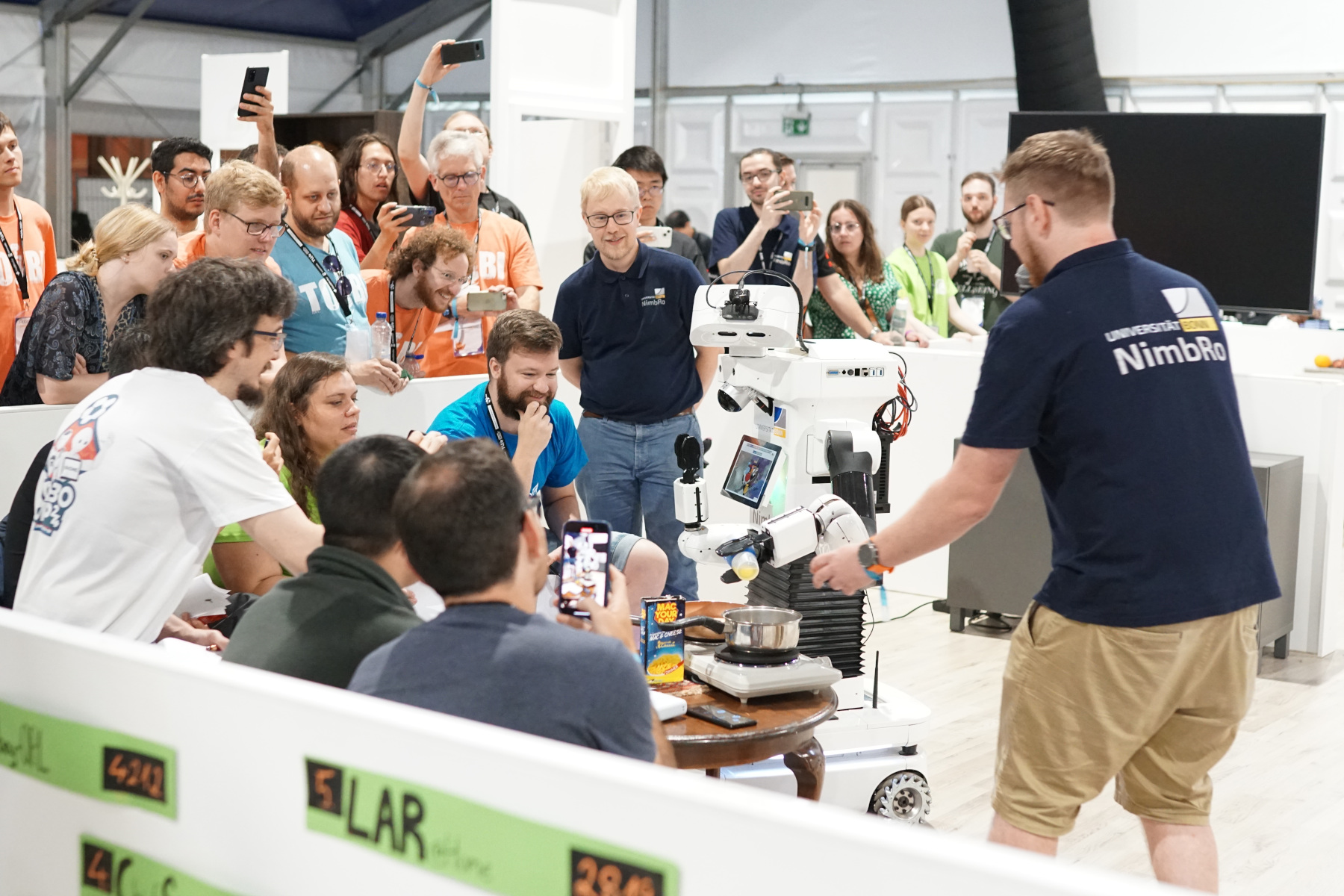}};
      \node[right=of a] (b) {\includegraphics[height=2.8cm]{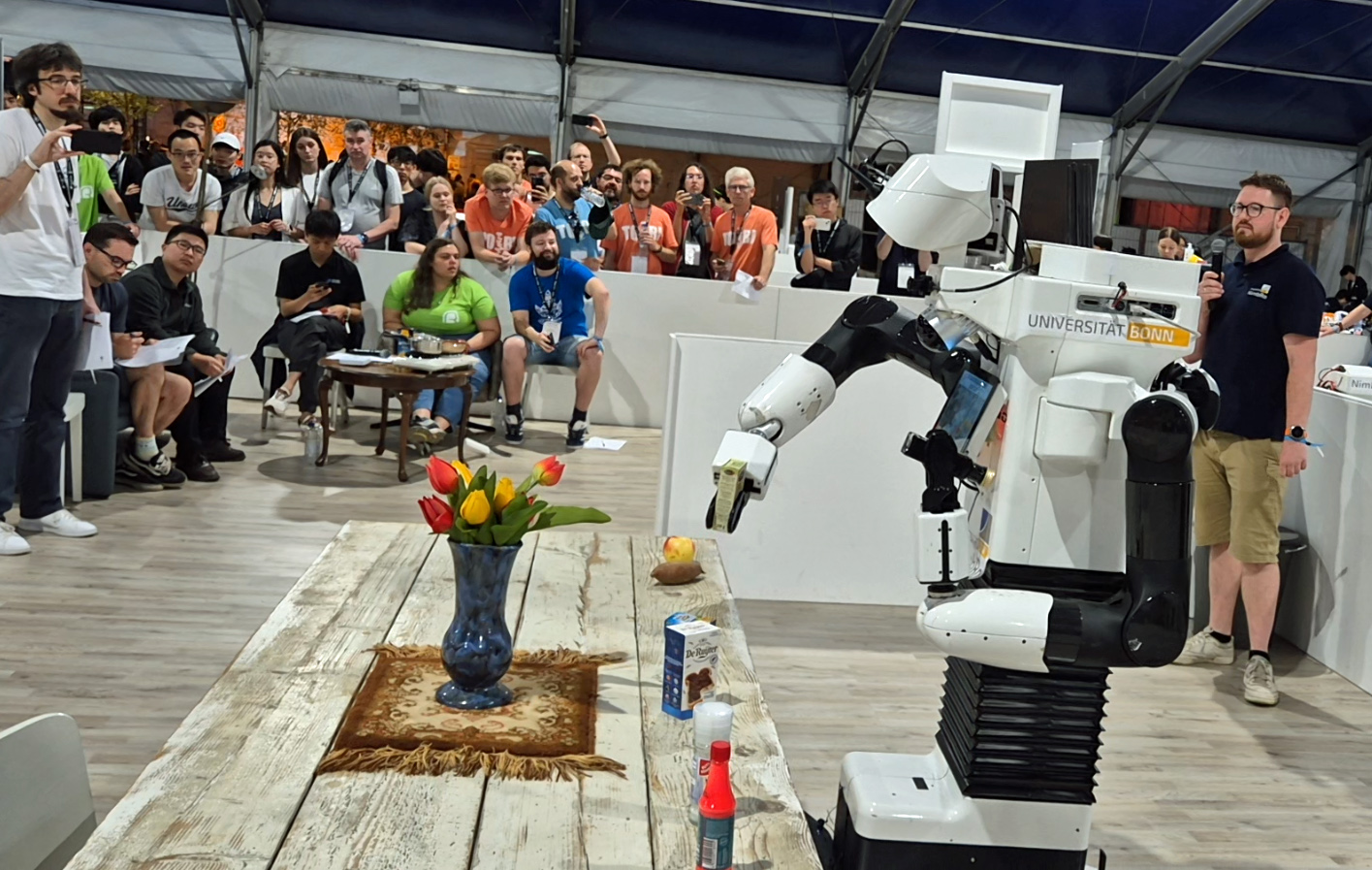}};
      \node[right=of b] (c) {\includegraphics[height=2.8cm, width=3.5cm]{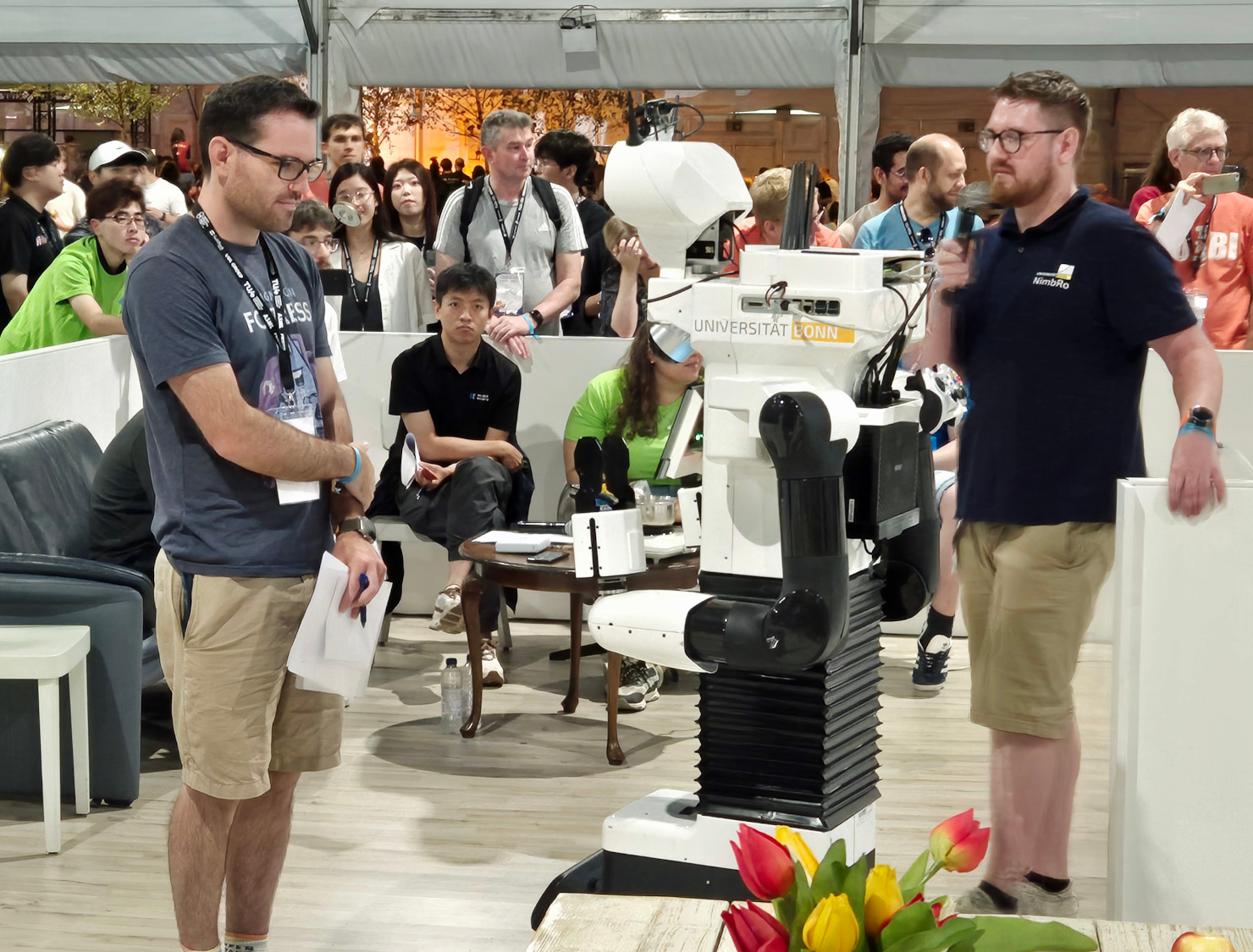}};
      \node[below=of a.south,align=center,yshift=+.05cm] (a_caption) {(a) Egg pouring};
      \node[below=of b.south,align=center,yshift=+.05cm] (b_caption) {(b) Apartment scanning};
      \node[below=of c.south,align=center,yshift=+.05cm] (c_caption) {(c) Human interaction};
    \end{tikzpicture}
    \vspace{-.8cm}
    \caption{Impressions from the final demonstration.}
    \vspace*{-3ex}
    \label{fig:finals}
\end{figure}
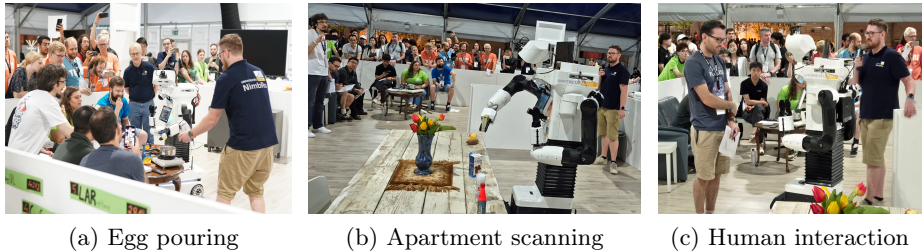

This year's theme for the final demonstration was "The robot helps a person in preparing dinner".
We aimed to showcase approaches and tasks that had not been shown in the predefined stages of the competition.
Due to the limited time of 10\,minutes available in the final demonstrations, we employed two robots in a time-splitting manner.
The first robot was used to scan the environment to find out which objects are present in the kitchen.
Next to a 2D map and marked locations, no further input was provided.
This part of our final demonstration should showcase how open-vocabulary vision approaches can be practically employed in domestic service robots to gather information about the environment and utilize this information for task planning based on user input.
The second robot was used to pour an egg into a pan.
This part should story-wise be after the scan of the environment to showcase that open-vocabulary approaches can be utilized to grasp non-labeled objects and allow achieving complex tasks like pouring an egg into a pan.
Impressions of our final demonstration are shown in \Cref{fig:finals}. 
For the final, our team received the highest scores from both the internal and the external jury members (1,783\,+\,2,152), followed by Tidyboy-OPL (1,545\,+\,1,738) and SocRob@Home (1,185\,+\,1,469).

\section{Lessons Learned}
\label{sec:lessons_learned}

Closed-set object detection approaches lack generalization in domestic service robotics. Open-vocabulary instance segmentation proved valuable in this competition. In the final demo, we used multimodal foundation models with open-vocabulary segmentation, enabling the robot to gather semantic scene information and grasp unseen objects. Capturing object relevance and attributes surpasses generic class label limitations. We successfully used a hybrid approach, combining closed-set detectors with open-vocabulary models for challenging categories like drinks, shoes, and garbage.
For future tasks, we aim to reduce reliance on custom closed-set models and enhance open-vocabulary segmentation integration with \ac{LLM} agents for arbitrary task planning in unseen environments.

Robustness and generalization are key for successful RoboCup@Home participation. We used hybrid Wi-Fi and 5G for connectivity, sensor watchdogs for reliability, and promptable object perception models for quick adaptation. 
Non-successful grasps were detected by reading the gripper's encoder states to reattempt the grasp. After too many failures, we triggered a handover procedure.

From last year's participation, we learned to prioritize natural interaction, emphasizing speech support and visual feedback on a touch screen interface. 
If speech commands are unclear, a graphical user interface allows users to select options or cancel phases. We also displayed object perception results for transparency and debugging.

\section{Conclusion}
\label{sec:conclusion}

In this paper, we summarized our hardware, approaches, and performance of the NimbRo@Home team at the RoboCup@Home 2024 competition in the Open Platform League.
For this year's participation, we successfully employed open-vocabulary approaches for object segmentation and \acp{LLM} for task planning and execution.
We successfully demonstrated that open-vocabulary approaches can be utilized to grasp non-labeled objects.
During the competition, we scored in all tasks of Stage~1 and Stage~2.
In the final demonstration, we showcased how open-vocabulary object perception approaches can be practically employed in domestic service robots to gather information about the environment and utilize this information for task planning based on user input and finally execute complex tasks like pouring an egg into a pan. 
Our team NimbRo@Home won the overall competition in the Open Platform League, followed by team Tidyboy-OPL (South Korea) and SocRob (Portugal).
Further, we received a \textit{Waitress Captain (Best in Restaurant Test)} award.
We believe that robustness and generalization are key for successful RoboCup@Home participation.


\begin{credits}
\subsubsection{\ackname} This work has been funded by the German Ministry of Education and Research, grant 16SV8683: Transferzentrum Roboter im Alltag (RimA).
\end{credits}

%
%
%
\bibliographystyle{splncs04}
\bibliography{references}

\begin{thebibliography}{10}
\providecommand{\url}[1]{\texttt{#1}}
\providecommand{\urlprefix}{URL }
\providecommand{\doi}[1]{https://doi.org/#1}

\bibitem{BeulNQRHPHB:JFR19}
Beul, M., Nieuwenhuisen, M., Quenzel, J., Rosu, R.A., et~al.: Team {NimbRo} at {MBZIRC} 2017: Fast landing on a moving target and treasure hunting with a team of micro aerial vehicles. Journal of Field Robotics (JFR)  \textbf{36}(1),  204--229 (2019)

\bibitem{BeulSQSBSRPRLSSSB:FR22}
Beul, M., Schwarz, M., Quenzel, J., Splietker, M., Bultmann, S., Schleich, D., Rochow, A., et~al.: Target chase, wall building, and fire fighting: Autonomous {UAVs} of team {NimbRo} at {MBZIRC} 2020. Field Robotics  \textbf{2}(1),  807--842 (2022)

\bibitem{bode2024prompting}
Bode, J., Pätzold, B., Memmesheimer, R., Behnke, S.: A comparison of prompt engineering techniques for task planning and execution in service robotics. In: IEEE-RAS 23rd International Conference on Humanoid Robots (Humanoids) (2024)

\bibitem{vad}
Bredin, H., Laurent, A.: End-to-end speaker segmentation for overlap-aware resegmentation. In: Interspeech Conference. pp. 3111--3115 (2021)

\bibitem{vgg-face}
Cao, Q., Shen, L., Xie, W., Parkhi, O.M., Zisserman, A.: {VGGFace2}: A dataset for recognising faces across pose and age. In: 13th IEEE International Conference on Automatic Face \& Gesture Recognition (FG). pp. 67--74 (2018)

\bibitem{JACK}
Davis, P., Letz, S.: \url{https://jackaudio.org} (2023)

\bibitem{Eren_Coqui_TTS_2021}
Eren, G., {Coqui TTS}: {Coqui TTS}. \url{https://github.com/coqui-ai/TTS} (2021)

\bibitem{rulebook_2024}
Hart, J., Moriarty, A., Pasternak, K., Kummert, J., Hawkin, A., Hassouna, V., Pena~Narvaez, J.D., et~al.: {RoboCup@Home} 2024: Rules and regulations (2024)

\bibitem{Jocher_Ultralytics_YOLO_2023}
Jocher, G., Chaurasia, A., Qiu, J.: {Ultralytics YOLOv8}. \url{https://github.com/ultralytics/ultralytics} (2023)

\bibitem{tts}
Kim, J., Kong, J., Son, J.: Conditional variational autoencoder with adversarial learning for end-to-end text-to-speech. In: 38th International Conference on Machine Learning (ICML). pp. 5530--5540 (2021)

\bibitem{kirillov2023segment}
Kirillov, A., Mintun, E., Ravi, N., Mao, H., Rolland, C., Gustafson, L., Xiao, T., Whitehead, S., Berg, A.C., Lo, W.Y., et~al.: Segment anything. In: IEEE/CVF International Conference on Computer Vision (ICCV). pp. 4015--4026 (2023)

\bibitem{LenzQPRRSSSSB:FR22}
Lenz, C., Quenzel, J., Periyasamy, A.S., Razlaw, J., Rochow, A., Splietker, M., Schreiber, M., Schwarz, M., et~al.: Autonomous wall-building and firefighting: Team {NimbRo's} {UGV} solution for {MBZIRC} 2020. Field Robotics  \textbf{2}(1),  55--74 (2022)

\bibitem{lenz2023nimbro}
Lenz, C., Schwarz, M., Rochow, A., P{\"a}tzold, B., et~al.: {NimbRo} wins {ANA Avatar XPRIZE} immersive telepresence competition: Human-centric evaluation and lessons learned. International Journal of Social Robotics (SORO)  (2023)

\bibitem{li2023mask}
Li, F., Zhang, H., Xu, H., Liu, S., Zhang, L., Ni, L.M., et~al.: {Mask DINO}: Towards a unified transformer-based framework for object detection and segmentation. In: IEEE/CVF Conf. on Computer Vision and Pattern Recognition (CVPR) (2023)

\bibitem{macenski2021slam}
Macenski, S., Jambrecic, I.: {SLAM Toolbox}: {SLAM} for the dynamic world. Journal of Open Source Software  \textbf{6}(61), ~2783 (2021)

\bibitem{macenski2022robot}
Macenski, S., Foote, T., Gerkey, B., et~al.: {Robot Operating System 2}: Design, architecture, and uses in the wild. Science robotics  \textbf{7}(66),  eabm6074 (2022)

\bibitem{memmesheimernimbrotdp2023}
Memmesheimer, R., Bode, J., Splietker, M., Bultmann, S., Imbusch, B.T., Behnke, S.: {NimbRo@Home} 2023 {Open} {Platform} {League} team description. RoboCup@Home Team Description Papers (2023)

\bibitem{Maxine}
NVIDIA: {NVIDIA MAXINE Audio Effects}. \url{https://github.com/NVIDIA/MAXINE-AFX-SDK} (2023)

\bibitem{openai2023gpt4}
OpenAI: {GPT-4} technical report (2023)

\bibitem{pages2016tiago}
Pages, J., Marchionni, L., Ferro, F.: {TIAGo}: The modular robot that adapts to different research needs. In: Int. WS on Robot Modularity, IROS. vol.~290 (2016)

\bibitem{Pavlichenko:Winner2024}
Pavlichenko, D., Ficht, G., et~al.: {RoboCup 2023 Humanoid AdultSize} winner {NimbRo}: {NimbRoNet3} visual perception and responsive gait with waveform in-walk kicks. In: RoboCup 2023: Robot World Cup XXVI. Springer (2024)

\bibitem{radford2023robust}
Radford, A., Kim, J.W., Xu, T., Brockman, G., McLeavey, C., Sutskever, I.: Robust speech recognition via large-scale weak supervision. In: International Conference on Machine Learning (ICML). pp. 28492--28518. PMLR (2023)

\bibitem{SchwarzBDSPLSB:Frontiers16}
Schwarz, M., Beul, M., Droeschel, D., Sch{\"{u}}ller, S., Periyasamy, A.S., Lenz, C., Schreiber, M., et~al.: Supervised autonomy for exploration and mobile manipulation in rough terrain with a centaur-like robot. Frontiers Robotics {AI}  \textbf{3}, ~57 (2016)

\bibitem{SchwarzDLPPRRSS:JFR19}
Schwarz, M., Droeschel, D., et~al.: Team {NimbRo} at {MBZIRC} 2017: Autonomous valve stem turning using a wrench. Journal of Field Robotics  \textbf{36}(1),  170--182 (2019)

\bibitem{SchwarzLGKPSB:ICRA18}
Schwarz, M., Lenz, C., Garc{\'{\i}}a, G.M., Koo, S., Periyasamy, A.S., Schreiber, M., Behnke, S.: Fast object learning and dual-arm coordination for cluttered stowing, picking, and packing. In: Int. Conf. on Robotics and Automation (ICRA) (2018)

\bibitem{schwarz2017nimbropicking}
Schwarz, M., Milan, A., Lenz, C., Munoz, A., Periyasamy, A.S., Schreiber, M., Sch{\"u}ller, S., Behnke, S.: {NimbRo Picking}: Versatile part handling for warehouse automation. In: IEEE Int. Conf. on Robotics and Automation (ICRA) (2017)

\bibitem{schwarz2017nimbro}
Schwarz, M., Rodehutskors, T., Droeschel, D., Beul, M., Schreiber, M., Araslanov, N., et~al.: {NimbRo} {Rescue}: solving disaster-response tasks with the mobile manipulation robot {Momaro}. Journal of Field Robotics (JFR)  \textbf{34}(2),  400--425 (2017)

\bibitem{serengil2024lightface}
Serengil, S., Ozpinar, A.: A benchmark of facial recognition pipelines and co-usability performances of modules. J. Information Techn.  \textbf{17}(2),  95--107 (2024)

\bibitem{StucklerSB:Frontiers16}
St{\"{u}}ckler, J., Schwarz, M., Behnke, S.: Mobile manipulation, tool use, and intuitive interaction for cognitive service robot {Cosero}. Frontiers Robotics {AI}  \textbf{3}, ~58 (2016)

\bibitem{sundaralingam2023curobo}
Sundaralingam, B., Hari, S.K.S., Fishman, A., Garrett, C., Van~Wyk, K., et~al.: {cuRobo}: Parallelized collision-free robot motion generation. In: IEEE International Conference on Robotics and Automation (ICRA). pp. 8112--8119 (2023)

\bibitem{fasterwhisper}
Systran: {Faster Whisper}. \url{https://github.com/SYSTRAN/faster-whisper} (2024)

\bibitem{DBLP:conf/fusion/WojkeMP17}
Wojke, N., Memmesheimer, R., Paulus, D.: Joint operator detection and tracking for person following from mobile platforms. In: 20th International Conference on Information Fusion (FUSION). {IEEE} (2017)

\bibitem{zhao2024open}
Zhao, X., Chen, Y., Xu, S., Li, X., Wang, X., Li, Y., Huang, H.: An open and comprehensive pipeline for unified object grounding and detection. arXiv preprint arXiv:2401.02361  (2024)

\end{thebibliography}
\end{document}